%% file: iclr2024_conference.tex
\documentclass{article}

\usepackage{arxiv}

\usepackage{subcaption}
\usepackage{algpseudocode}
\usepackage{dsfont}
\usepackage{natbib}
\usepackage[dvipsnames]{xcolor}

\usepackage{url}
\usepackage{amsmath}
\usepackage{amssymb}
\usepackage{mathtools}
\usepackage{amsthm}
\usepackage{multirow,tabularx}
\usepackage{longtable}
\usepackage{booktabs,longtable}
\usepackage{hhline}
\usepackage{microtype}
\usepackage{graphicx}
\usepackage{booktabs}
\usepackage{algorithm}
\usepackage{pifont}
\usepackage{enumitem}
\usepackage{tcolorbox}
\usepackage{wrapfig}
\usepackage{amsmath}
\usepackage{bm}
\usepackage{fontawesome}

\definecolor{mg}{RGB}{0, 210, 140}
\definecolor{mr}{RGB}{251, 111, 111}
\definecolor{star}{RGB}{0, 147, 0}
\definecolor{mb}{RGB}{142, 156, 255}
\definecolor{mpink}{RGB}{242, 215, 151}
\definecolor{mpurple}{RGB}{216, 176, 255}

\newcommand{\ours}{\texttt{Semantic-Cohesive Knowledge Distillation for Deep Cross-Modal
Hashing}}

\usepackage{fancyhdr} 
\usepackage[font={footnotesize}]{caption}
\usepackage{hyperref}

\hypersetup{
 colorlinks=true,
 citecolor=blue,
 linkcolor=black,
 urlcolor=black}

\usepackage{cleveref}

\title{
Semantic-Cohesive Knowledge Distillation for Deep Cross-modal Hashing 
}

\author{Changchang Sun$^{1}$ 
~~~~Vickie Chen$^2$  
~~~~Yan Yan$^1$\\
  $^1$University of Illinois Chicago
  ~~~~~ $^2$Rensselaer Polytechnic Institute
  \\
  {\tt\small \{csun47, yyan55\}@uic.edu}
  ~~~~~ {\tt\small chenv4@rpi.edu}
}

\date{}

\begin{document}
\maketitle

\begin{abstract}
Recently,
deep supervised cross-modal hashing methods have achieve compelling success by learning semantic information in a self-supervised way.
 However, 
 they still suffer from the
 key 
 limitation 
 that
 the multi-label semantic extraction process
fail to
 explicitly interact with raw multimodal data,
making the learned representation-level semantic information not compatible with the heterogeneous multimodal data and hindering the performance of bridging modality gap. 
 To address this limitation,
in this paper,
we propose a novel semantic cohesive knowledge distillation scheme for deep cross-modal hashing, dubbed as SODA. 
Specifically, the multi-label information is introduced as a new textual modality and reformulated as a set of ground-truth label
prompt, depicting the semantics presented in the image like the text modality. 
Then,
a cross-modal teacher network is devised to effectively 
distill cross-modal semantic characteristics between image and label modalities
and thus learn a 
well-mapped Hamming space for image modality.
In a sense, 
such Hamming space can 
be regarded as a kind of prior knowledge to guide the learning of cross-modal student network and
comprehensively
preserve the semantic similarities between image and text modality.
Extensive experiments on two benchmark datasets demonstrate the superiority of our model over the state-of-the-art methods.
\end{abstract}

\input{1-intro}

\input{2-rel}

\input{3-model}

\input{4-exp}

\input{5-con}


    \bibliographystyle{iclr2024_conference}
    \bibliography{iclr2024_conference}



\end{document}

%% file: 1-intro.tex
\section{Introduction}
With the unprecedented growth of multimedia data on the Internet, 
cross-media retrieval which aims to search
semantically similar instances in one modality (\textit{e.g.}, image)
with a query of another modality (\textit{e.g.}, text) have become a compelling research topic recently.
 Due to the remarkable advantages of fast retrieval speed and low storage cost,
 cross-modal hashing methods~\citep{ZhouDG14,WangLWZZ15,MoranL15,LuZCSZ19,MandalCB19,WuWS19,ChenZMLY18,MaLMWN18,DongNLGW18,LiuLDPF18,CaoLLW18} that map the heterogeneous high-dimensional multimodal data from original space to a common Hamming space with limited hash code bits
 have 
gained a surge of research interest. 
Essentially, 
the major concern of cross-modal hashing methods is to 
preserve the inter-modal semantic similarity
and generate similar hash codes for semantically relevant instances.
According to the utilization of category label information,
existing cross-modal hashing methods can be roughly divided into two groups: unsupervised methods~\citep{DingGZ14,ZhuHSZ13,MasciBBS14,IrieAT15,ZhangZHT15,SongYYHS13,ZhouDG14,DingGZG16} and supervised ones~\citep{JiangL17,YuWYTLZ14,SunSFZZN19,ZhenHWP19,ChenWLLNX19,LiDL0GT18,DengCLGT18,ZhangL14}.
Benefiting from the advantages of exploring semantic labels
to 
guide the cross-modal hashing learning,
increasing efforts have been dedicated to the supervised manner.

\begin{wrapfigure}{r}{0.53\textwidth}
\centering
\vspace*{-4mm}
\begin{subfigure}[t]{0.46\linewidth}
    \centering
    \includegraphics[width=\linewidth]{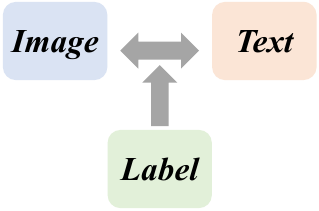}
    \caption{Pairwise Oriented}
    \label{pattern1}
\end{subfigure}
\hfill
\begin{subfigure}[t]{0.46\linewidth}
    \centering
    \includegraphics[width=\linewidth]{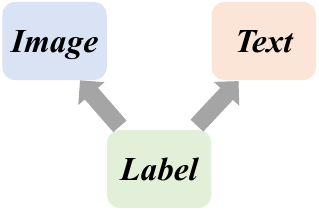}
    \caption{Self-supervised Oriented}
    \label{pattern2}
\end{subfigure}
\vspace*{-1mm}
\caption{
Illustration of two model optimization strategies of existing cross-modal hashing methods.
}
\label{cross}
\vspace*{-6mm}
\end{wrapfigure}

In fact, based on the role of category label information played in the 
hash code learning procedure,
existing supervised cross-modal hashing efforts
have two classic and representative optimization strategies.
Here, we name them as pairwise oriented and self-supervised oriented.
Specifically,
in terms of ``pairwise oriented" line,
early studies~\citep{JiangL17,SunSFZZN19} mainly focus on leveraging the 
pairwise similarity matrix constructed 
according to the label vectors 
to
guide the cross-modal hashing learning between image and text modalities,
as shown in Fig.~\ref{pattern1}.
However, in many  real-world scenarios, instances are often annotated with multi-labels, like the mainstream cross-modal datasets MIRFLICKR-25K~\citep{HuiskesL08} and NUS-WIDE~\citep{ChuaTHLLZ09}.
It is thus inappropriate to 
 measure the semantic similarity among instances simply by
counting their common labels and
neglect rich semantic information contained in multi-labels. 
Therefore, to address this issue, 
some methods such as ~\citet{LiDL0GT18} expect to first 
extract 
representation-level semantic information from 
one-hot multi-label vectors
in a self-supervised manner and learn a label Hamming space.
Then, the hash code learning processes of each modality 
is guided by 
utilizing the learned
label Hamming space,
 as shown in Fig.~\ref{pattern2}.
In a sense,
the learned semantic information from multi-labels
acts as an intermediate bridge and 
enforces the hash code learning of image and text modalities fitting to the pre-learned label Hamming space.

Although existing ``self-supervised oriented" cross-modal hashing optimization strategy has achieved compelling success regarding multi-label data, it still suffers from three critical limitations:
(i)  
Due to the fact that the number of category labels of one dataset is specific,
the semantic information can be learned from the one-hot label vectors is limited. 
Accordingly,
cross-modal hashing learning performance will be sub-optimal if these pre-defined label features are not well characterized.
Besides,
it is worth noting that heterogeneous cross-modal data 
contains rich and
complicated characteristics, such as the color and texture features of images and the semantic information of text description.
In the light of this, 
existing studies~\citep{LiDL0GT18} that only enforce the cross-modal data mapping into a pre-defined Hamming space learned from one-hot label vectors 
will overlook the rich semantic features of origin cross-modal data. 
The feature extraction backbones of image and text will be more inclined to realize the best match with the pre-defined semantically impoverished
label Hamming space and limited to 
truly
exploit rich semantic features from original data, causing poor retrieval performance in the testing phase.
(ii)
For most existing cross-modal hashing methods, the text and label modalities are represented as the one-hot vector based on the bag-of-words (BoW) strategy, where rich semantic information conveyed by the
text and label description is ignored. 
(iii)
Learning representation-level semantic information in a self-supervised manner neglects the explicit cross-modal feature interaction.
On the one hand, the explicit feature aliments
between image and label modalities
are overlooked when generating the label Hamming space, making the semantic representation learned from label modality not compatible with the
feature distribution of the image modalities.
On the other hand,
the text description of two images that have same category labels may vary greatly. 
Therefore,
existing methods that directly perform text-label feature aliment may reach sub-optimal results, failing to acquire similar hash codes for truly semantically similar text instances.

\begin{figure*}[htp]
	\centering
\setlength{\abovecaptionskip}{0.3cm}
	\includegraphics[scale=0.52]{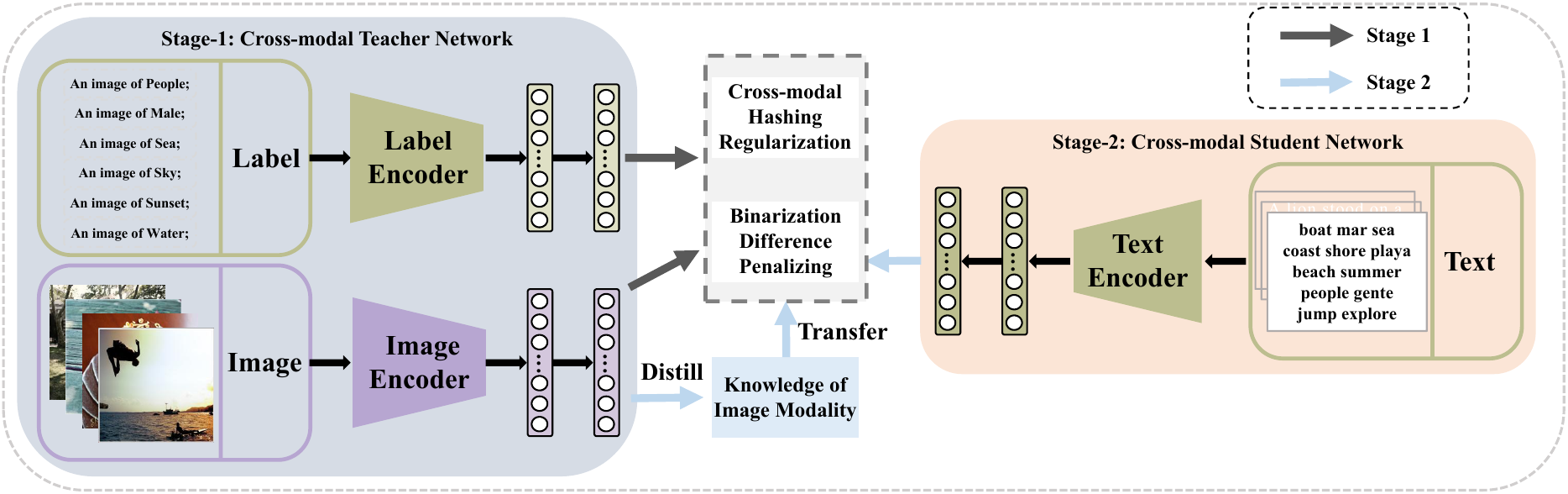}
	\vspace{-3ex}
	\caption{Illustration of the proposed SODA, where the cross-modal teacher network is first designed to distill knowledge of image modality by directly narrowing the image-label modality gap and the cross-modal student network is then trained using the distilled image modality knowledge. 
}\label{framework}
     \vspace{-0.3cm}
\end{figure*}

To address aforementioned limitations, as shown in Fig.~\ref{framework}, we propose
a novel 
semantic-cohesive knowledge distillation scheme for
cross-modal hashing learning, SODA for short,  
where a two-stage cross-modal teacher-student network is devised
to totally capture the cross-modal semantic characteristics between different modalities with knowledge propagating.
Specifically, the multi-label information is introduced as a new textual modality, depicting the semantic elements presented in the image in a more intuitive way. 
To reformulate
ground-truth multi-labels as integrated text, 
we resort to the 
prompt engineering~\citep{BrownMRSKDNSSAA20} and characterize 
the category labels of each instance with a set of ground-truth label prompt.
Besides, 
motivated by the fact that, compared with the text modality,  
the label modality is more discriminative and 
suitable to capture the common semantic characteristics of semantically similar images. 
We thus 
first devise a cross-modal teacher network to maximize the semantic relevance and the feature distribution consistency between image and label modalities.
In this way, the image modality and label modality can be well mapped into a common Hamming space with the cross-modal similarity correlation preserving.
Thereafter, 
based on the learned
common Hamming space 
regarding image and label modalities,
the hash code learning of the text modality can be effectively performed by
fitting the image modality.
Here, even though the text descriptions of two semantically similar images differ greatly, they can be well optimized under the supervision of 
well-learned image Hamming space.

Our main contributions can be summarized in threefold: 
\sloppy

$\bullet$ To the best of our knowledge, this is the first attempt to tackle the problem of supervised cross-modal hashing 
using a teacher-student optimization strategy by
propagating cross-modal knowledge learned from image and label modalities to guide the hash code learning of the text modality.

$\bullet$ We design 
 a image-label teacher network to learn the discriminative image Hamming space by mutually narrowing the gap between image and label modalities,
 which can be seamlessly adopted as the knowledge to regularize the hash learning of the text modality in the following image-text student network.
 
$\bullet$ 
 We present category labels using ground-truth label prompt set and directly interact with image modality,
 solving the problem that the learned semantic features are not compatible with the target cross-modal data.
 Extensive experiments demonstrate the superiority of SODA over the state-of-the-art methods on two benchmark datasets. 

%% file: 2-rel.tex
\section{Related Work}
In this section,
the most related methods on the topic of unsupervised and supervised cross-modal hashing methods will be reviewed and elaborated one by one.

\subsection{Unsupervised Cross-modal Hashing}
Unsupervised cross-modal hashing methods~\citep{DingGZ14,GongL11,WuLHLDZS18,LiuQGZY20,YuZZT21} aim to learn the hash mapping function 
and bridge the modality heterogeneity gap
based on the correlation information naturally existing in the paired cross-modal data.
For instance, to learn the unified hash codes, \citet{DingGZ14} resorted to the collective matrix factorization with latent factor model from different modalities of one instance, and hence improved the cross-modal search accuracy by merging multiple view information.
However, such matrix factorization based methods 
suffer from the inferior relaxation strategy, where the 
discrete constraints  are discarded when learning the hash function.
Therefore, to address this issue, \citet{WuLHLDZS18} presented a unsupervised deep learning framework, where the deep learning and matrix factorization are jointly integrated in a self-taught manner.
Besides, by utilizing the original neighborhood relations from different modalities, \citet{SuZZ19} devised a joint-semantics affinity matrix to further capture the latent intrinsic semantic affinity of the multi-modal instances.
In addition,
to fully preserve the semantic correlations among
instances and enhance the discriminative ability of learned
hash codes, \citet{LiuQGZY20}
proposed a 
novel joint-modal distribution-based similarity hashing method and introduced a better sampling and weighting scheme.
Overall, although existing unsupervised methods have achieved
compelling performance, they suffer from the limitations of the lack of representation-level supervision and hence 
cannot meet the requirements of retrieval accuracy in the real-world applications.

\subsection{Supervised Cross-modal Hashing }
In contrast, 
supervised cross-modal hashing methods~\citep{ZhangL14,YuWYTLZ14,LinDHW17,JiangL17,LiDL0GT18,RafailidisC16,ZhangZLG14,ZhangLLNX17,0002WLCZH21,ShenLYXHSH21} work on leveraging semantic labels 
as the supervision 
to explicitly guide hash codes learning.
In this way, the similarity relationship in the original data space can be well preserved in the Hamming space and hence boost the cross-modal retrieval performance.
Generally, the semantic labels are utilized to construct a binary similarity matrix or
establish a cross-modal optimization goal to minimizing the modality difference.
For example, 
to seamlessly integrate semantic labels into the hashing learning procedure for large-scale data modeling, \citet{ZhangL14}
introduced 
a sequential learning method to learn the hash
functions bit by bit with linear-time complexity.
Besides, \citet{XuSYSL17} proposed a novel discrete cross-modal hashing method, where the discriminability
of labels can be explicitly captured and the online
retrieval time is largely reduced.
Inspired by the remarkable representation capacity of deep neural networks, \citet{JiangL17} established the first end-to-end cross-modal deep hashing framework to perform the feature learning from scratch. 
Due to the fact that it is pretty time-consuming and knowledge-required to annotate a large amount of dataset, the real application of supervised cross-modal hashing is largely limited.
Therefore, \citet{HuXH020} focused on the idea of knowledge distillation,
where the similarity relationships are first distilled using outputs produced by an unsupervised method and then a supervised model is 
efficiently
optimized under the guidance of such semantic information.
In addition,
 due to the concern that low-quality annotations inevitably introduce numerous mistakes, \citet{YangYLD22} designed a robust cross-modal hashing framework to correlate distinct
modalities and combat noisy labels simultaneously.
Furthermore, 
to characterize the latent structures that exist among different modalities, \citet{0002WLCZH21} proposed a graph
 convolutional networks (GCNs) to exploit the local structure information of datasets for cross-modal hash learning.

Beyond that,
to better take advantage of the multiple category labels and describe the semantic relevance across different modalities more accurately,
many methods also target at first learning a semantic representation from multi-label inputs directly, and then supervise the hash learning processes utilizing the learned semantic features.
For example, \citet{LiDL0GT18} designed a self-supervised adversarial hashing method, where the high-dimensional features and their corresponding hash codes of different modalities are jointly characterized under the guidance of the learned semantic subspace.
However, 
although these methods have achieved compelling performance,
they suffer from the limitation that the representation-level semantic supervision is obtained in a self-supervised manner and directly taken as the cross-modal hashing optimization target.
In fact, the learned semantic representation may not be fully compatible with the heterogeneous cross-modal data and hence 
result in inferior
performance.
Towards this end, in our work, 
to eliminate the modality gap,
we design a semantic-cohesive knowledge distillation method for deep cross-modal hashing.

%% file: 3-model.tex
\section{Preliminaries}
We first introduce the necessary notations throughout the paper, and then define the studied task.
\paragraph{Notations.} To simplify the presentation, we focus on the cross-modal retrieval for the bimodal data (\textit{i.e.}, the image and text). Without losing the generality, our task can be easily extended to the scenarios with other modalities.
Suppose that we have $N$ multi-labeled instances $\mathcal{E}{=}\left\lbrace e_{i}\right\rbrace_{i=1}^{N}$, where $e_i$ refers to the $i$-th instance. Each instance is comprised of an image, a text description, and a category label set, \textit{i.e.}, $e_i {=}({v}_{i}, {t}_{i}, {y}_{i})$.
In particular, if instance $e_{i}$ is labeled with 
a series of $K$ categories $y_i{=}\{y_i^1,y_i^2,\cdots,y_i^K\}$, we resort to the prompt engineering~\citep{BrownMRSKDNSSAA20} and
design the prompt by posing a blank-filling problem for each category. For example, for the $k$-th category, the prompt format is ``$\textit{An image of}$ $y_i^k$".
 Moreover, according to the category labels, 
 we also introduce two binary cross-modal similarity matrices $\mathbf{S}^{tea}$ and $\mathbf{S}^{stu}$
to
globally 
determine whether two instances are similar or not, where $\mathbf{S}^{tea}_{ij}{=}1$ if image $v_i$ has at least one category belonging to $y_j$, and $\mathbf{S}^{tea}_{ij}{=}0$ otherwise. In a similar manner, 
 $\mathbf{S}^{stu}_{ij}{=}1$ if image $v_i$ shares at least one common category with $t_j$, and  $\mathbf{S}^{stu}_{ij}{=}0$ otherwise.
\paragraph{Problem Formulation.}
In this work, we aim to devise a novel two-stage teacher-student cross-modal hashing network to obtain the accurate $L$-bit hash codes of each modality for the $i$-th instance, namely, $\mathbf{b}_{v_i}\in{\left\lbrace-1,1\right\rbrace }^{L}$, $\mathbf{b}_{t_i}\in{\left\lbrace -1,1\right\rbrace }^{L}$, and
$\mathbf{b}_{y_i}\in{\left\lbrace -1,1\right\rbrace }^{L}$.
Based on the hash codes, we can measure the inter-modal similarities using the Hamming distance as $dis_H(\mathbf{b}_{v_i},\mathbf{b}_{t_j}){=}\frac{1}{2}(L-\mathbf{b}_{v_i}^T\mathbf{b}_{t_j})$ and hence perform the cross-modal retrieval.
Specifically, 
the hash code learning process of each modality can be denoted as 
$ \mathbf{b}_{v_i}{=}sgn(f^{v}({v}_{i};\mathbf{\Theta}_{v}))$, $ \mathbf{b}_{t_i}{=}sgn(f^{t}({t}_{i};\mathbf{\Theta}_{t}))$), and
$ \mathbf{b}_{y_i}{=}sgn(f^{y}({y}_{i};\mathbf{\Theta}_{y}))$), respectively.
$sgn(\cdot)$ is the element-wise sign function, which outputs $``+1"$ for positive real numbers and $``-1"$ for negative ones. Here, $f^{v}$, $f^{t}$ and $f^{y}$ refer to the hashing networks with parameters $\mathbf{\Theta}_{v} $, $ \mathbf{\Theta}_{t}$ and $\mathbf{\Theta}_{y} $ to be learned.

\section{The Proposed Model}
In this section, we present the proposed SODA, as the major novelty, which is able to effectively leverage the 
image modality knowledge learned from the cross-modal teacher network to supervise the hash code learning of the text modality.
In particular, we first set up 
\emph{hash representation learning} to extract semantic features for each modality.
And then we introduce
\emph{cross-modal semantic knowledge distillation} 
to maximize the semantic relevance and the feature distribution consistency between
image and label modalities.
Last,
taking the learned image hash codes as an optimization medium,
\emph{cross-modal semantic supervision} is devised to learn the hash codes of the text modality by fitting the established image Hamming space.

\subsection{Hash Representation Learning}
Motivated by the 
strong representation capacity of the 
multimodal pre-training model CLIP~\citep{RadfordKHRGASAM21}, 
we resort to its image encoder and text encoder to perform feature extraction.
Concretely, regarding the image modality,
we initialize CLIP image encoder with the released base version consisting of $16$ transformer layers, followed by some fully connected neural networks to realize dimension reduction.
In particular, 
given $i$-th instance, 
we obtain the image hash representation $\mathbf{h}_{v_i}{=}f^{v}({v}_{i};\mathbf{\Theta}_{v})\in \mathbb{R}^{L}$.
As for the label and text modalities, similar with image modality, we integrate the CLIP text encoder with some fully connected layers two times, and input the constructed ground-truth label prompt set and original text description to obtain their hash representations, separately.
Formally,  $\mathbf{h}_{t_i}{=}f^{t}({t}_{i};\mathbf{\Theta}_{t})\in \mathbb{R}^{L}$ and
 $\mathbf{h}_{y_i}{=}f^{y}({y}_{i};\mathbf{\Theta}_{y})\in \mathbb{R}^{L}$.

\subsection{Cross-modal Semantic Knowledge Distillation}
To address the issue that the semantic information learned from multi-labels in a self-supervised way is not compatible with the image and text modalities,
we employ the regularization between image and label modalities to comprehensively preserve the semantic similarity in a ``cross-modal oriented" manner.
Specifically,
we design a cross-modal teacher network
and map
the image and label modalities into a common Hamming space.
In this way, the hash code learning of image modality can be realized under the supervision of label modality, which is more discriminative compared with the text modality.
In detail, 
we maximize the Hamming distance between two instances of image and text modalities whose semantic similarity is $0$, while minimizing that with the similarity of $1$.
We define the cross-modal semantic similarity in teacher network
using the continuous surrogates of the binary hash codes $\mathbf{h}_{v_i}$ and $\mathbf{h}_{y_j}$
as follows,
\begin{equation}
\phi^{tea}_{ij}=
\begin{aligned}
\dfrac{1}{2}(\mathbf{h}_{v_{i}})^{T}\mathbf{h}_{y_j},
\end{aligned}
\end{equation}
where $\phi^{tea}_{ij}$ denotes the semantic similarity between image and label instances.

Similar to~\cite{JiangL17}, we encourage $\phi^{tea}_{ij}$ to approximate the binary ground truth ${S}_{ij}^{tea}$ and obtain the cross-modal hashing  regularization component as follows,
\begin{equation}
L(\phi^{tea}_{ij} |S_{ij}^{tea})=\sigma(\phi_{ij}^{tea})^{S_{ij}^{tea}}(1-\sigma(\phi_{ij}^{tea}))^{(1-S_{ij}^{tea})},
\end{equation}
where $ \sigma(\cdot)$ is the sigmoid function. Simple algebra computations enable us to reach the following objective function,
\begin{equation}
\Phi_{1}=
\begin{aligned}
-\sum_{i,j=1}^{N}(S^{tea}_{ij}\phi^{tea}_{ij}-\log(1+e^{\phi^{tea}_{ij}})).
\end{aligned}
\end{equation} 
Meanwhile, a binarization difference penalizing~\citep{SunSFZZN19} is adopted to derive more powerful hash representations by
minimizing the difference between learned hash representation and hash codes.
The binarization difference regularization can be written as follows,
\begin{equation}
\Phi_{2}=
\begin{aligned}
\sum_{i,j=1}^{N}( \begin{Vmatrix}\mathbf{b}_{v_i}-\mathbf{h}_{v_i} \end{Vmatrix}_{F}^{2}+\begin{Vmatrix}\mathbf{b}_{y_j}-\mathbf{h}_{y_j} \end{Vmatrix}_{F}^{2}  ),
\end{aligned}
\end{equation}
where $\begin{Vmatrix}\cdot\end{Vmatrix}_F$ denotes the Frobenius norm.
Notably, to bridge the semantic gap between different modalities more effectively and boost the performance of the cross-modal hashing, we adopt the unified binary hash codes (\textit{i.e.}, $\mathbf{b}^{tea}_{i}{=}\mathbf{b}_{v_i}{=}\mathbf{b}_{y_i}$) in the training procedure. Towards this end, we have,
\begin{equation}
\mathbf{b}^{tea}_{i}=sgn\left(
\begin{aligned}
\mathbf{b}_{v_i}+ \mathbf{b}_{y_i}
\end{aligned}
\right).
\end{equation}
Consequently, we have the following objective function towards the cross-modal hashing learning between image and text modalities,
\begin{equation}
\begin{split}
\Psi_{tea}=&-\min \limits_{\mathbf{\Theta}_{v}, \mathbf{\Theta}_{y}}\sum_{i,j=1}^{N}\Big(S^{tea}_{ij}\phi^{tea}_{ij}-\log(1+e^{\phi^{tea}_{ij}})\Big)\\
&+\alpha( \begin{Vmatrix}\mathbf{b}_{i}^{tea}-\mathbf{h}_{v_i} \end{Vmatrix}_{F}^{2}+\begin{Vmatrix}\mathbf{b}_{j}^{tea}-\mathbf{h}_{y_j} \end{Vmatrix}_{F}^{2}),
\end{split}
\label{alpha}
\end{equation}
where $\alpha$ is the nonnegative tradeoff parameter.

\subsection{Cross-modal Semantic Supervision}
Having obtained the hash codes of image modality, 
the hash codes of the text modality can also be learned by taking the learned image hash codes as the prior knowledge.
In a sense,  
to preserve the similarity correlation between image and text modalities, 
we can learn 
the hash codes of the text modality by mapping it to the well-learned common Hamming space of image and label modalities.
 Towards this end, 
 similar with the learning of the cross-modal teacher network, the cross-modal student network can also be trained using the following objective function,
\begin{equation}
\begin{split}
\Psi_{stu}=&-\min \limits_{\mathbf{\Theta}_{t}}\sum_{i,j=1}^{N}\Big(S^{stu}_{ij}\phi^{stu}_{ij}-\log(1+e^{\phi^{stu}_{ij}})\Big)\\
&+\beta( \begin{Vmatrix}\mathbf{b}_{i}^{stu}-\mathbf{h}_{v_i} \end{Vmatrix}_{F}^{2}+\begin{Vmatrix}\mathbf{b}_{j}^{stu}-\mathbf{h}_{t_j} \end{Vmatrix}_{F}^{2}),
\end{split}
\label{beta}
\end{equation}
where $\beta$ is the nonnegative tradeoff parameter and 
$\phi^{stu}_{ij}$ can be written as follows,
\begin{equation}
\phi^{stu}_{ij}=
\begin{aligned}
\dfrac{1}{2}(\mathbf{h}_{v_{i}})^{T}\mathbf{h}_{t_j}.
\end{aligned}
\end{equation}
Similarly, $\mathbf{b}_{i}^{stu}$ can be obtained as follows,
\begin{equation}
\mathbf{b}^{stu}_{i}=sgn\left(
\begin{aligned}
\mathbf{b}_{v_i}+ \mathbf{b}_{t_i}
\end{aligned}
\right).
\end{equation}
Notably, the hash code of images are fixed in the student cross-modal network and act as the optimization medium of the text modality.

%% file: 4-exp.tex
\section{Experiments}
In this section, we present extensive experimental results and analysis on two datasets.

\subsection{Datasets}
For the evaluation, we adopted two widely used cross-modal benchmark datasets: MIRFLICKR-25K~\citep{HuiskesL08} and NUS-WIDE~\citep{ChuaTHLLZ09}, where images are assigned to multiple category labels.

\paragraph{MIRFLICKR-25K.}
This dataset includes $25,000$ images with the fixed size of $224{\times}224{\times}3$, which are  
originally collected from the Flickr website\footnote{http://www.flickr.com/.}. And each image is manually annotated with several textual tags and at least one of the $24$ labels.
In our experiments, we merely utilized images that are associated with at least 20 textual tags. Therefore, there are $20,015$ images retained.  
Afterwards, we split these images into two subsets: query and gallery. Specifically, $2,000$ images are randomly selected as the query subset, and the remaining ones are set as gallery set. To learn the hash function, $10,000$ images are randomly chosen from the gallery subset as training data. 
Moreover, to reduce noisy tags, we removed tags that appear below $20$ from retained images, and hence obtained $1,386$ unique tags.

 \begin{wraptable}{r}{0.32\textwidth}
\vspace{-0.14in} 
\centering
   \scalebox{0.7}{
   \begin{tabular}{p{1.7cm}<{\centering}|*{1}{p{2.7cm}<{\centering}|p{1.8cm}<{\centering}}}
		\hline
			&MIRFLICKR-25K&NUS-WIDE \\\hline 
				Query Set& $2,000$& $2,100$\\
		Training Set& $10,000$& $10,500$\\
		Gallery Set& $18,015$& $193,734$\\
	    Tags& $1,386$& $1,000$\\
		Labels& $24$& $21$\\
		\hline
	\end{tabular}
} 
\caption{Summary of the MIRFLICKR-25K and NUS-WIDE dataset used in our experiments. }\label{datasets}
\end{wraptable}
\paragraph{NUS-WIDE.}
It is a large-scale social image dataset
including $269,648$ images associated with $5,018$ unique tags, where
the image size is $224{\times}224{\times}3$. Moreover, each image is 
manually annotated by a predefined set of 81 labels. 
In our work, we retained $195,834$ images
that are associated with at least one of the $21$ most frequent labels.
Meanwhile, similar to MIRFLICKR-25K, 
we formed a query set of $2,100$ images, 
while the training set and gallery set containing $10,500$ and $193,734$ images, respectively.
And we removed those tags that appear below $20$ to construct the word bag and obtained 
$1,000$ unique tags. 
The statistics of datasets are summarized in Tab.~\ref{datasets}.


\subsection{Experimental Settings}
\paragraph{Evaluation Protocols.}
In this work, we evaluated our proposed model on two classic cross-modal retrieval tasks: querying the image database with given textual vectors (``Text$\rightarrow$Image'') and querying the text database with given image examples (``Image$\rightarrow$Text''). 
For each cross-modal retrieval task, we adopted two widely utilized performance metrics, \textit{i.e.}, Hamming ranking and hash lookup, to compare the retrieval performance of our method with other 
state-of-the-art methods. In particular, mean average precision (MAP)~\citep{XuSYSL17}, a representative method to measure the accuracy of Hamming ranking, is adopted in our work. 
Meanwhile, the precision-recall (P-R) curve is utilized to measure the accuracy of hash lookup protocol.
Notably, to be consistent with baseline methods,
two instances are considered to be similar if and only if they share at least one label in the testing phase.

\paragraph{Baselines.} To justify the effectiveness of our proposed SODA, we chose six state-of-the-art cross-modal hashing methods as baselines, including five supervised methods: SCM~\citep{ZhangL14}, DCH~\citep{XuSYSL17}, DCMH~\citep{JiangL17},  SSAH~\citep{LiDL0GT18}, and TECH~\citep{ChenWLLNX19}, and one unsupervised one: CCA~\citep{GongL11}. 
As SCM presents two learning models, \textit{i.e.}, orthogonal projection and sequential one, we
respectively denoted them by SCM-Or and SCM-Se.
Among these baselines, CCA, SCM-Or, SCM-Se, DCH, and TECH are shallow learning methods, namely they rely on hand-crafted image features. 
In our work, we adopted the image encoder and text encoder of pre-trained CLIP model.
For fairness, we separately extracted image features 
and text features from the same CLIP encoders for shallow learning approaches.    
Besides, we did not change the backbone of DCMH and SSAH as they cannot converge during training using the CLIP encoders.
Besides, in order to be consistent with existing methods and 
avoid the impact of feature extraction mode on retrieval performance, we also extracted $1,000$-d images features from CNN-F~\citep{ChatfieldSVZ14} networks that are pre-trained on Imagenet~\citep{DengDSLL009} for shallow learning methods. Meanwhile, based on the BoW strategy, the textual modality of each instance in MIRFLICKR-25K is represented by a $1,386$-d vector, and that in NUS-WIDE is represented as a $1,000$-d vector.
Note that the dataset partitioning of baseline methods is different and
the source codes and involved parameters of above baselines are kindly provided by corresponding authors, we hence re-run each baseline method using the unified data partitioning.
Besides, we tried our best to 
tune the models and reported their best performance as that in their papers.

\paragraph{Implementation Details.}
We implemented SODA with the open source deep learning software library PyTorch, 
and adopted the
adaptive moment estimation (Adam) gradient descent
as the optimizer~\citep{KingmaB14}.
The learning late is chosen from $10^{-6}$
to $10^{-8}$.
The image and text encoders are initialized with the base version of CLIP composing of $16$ layers, while other parameters are initialized randomly. 
To determine hyper-parameters, \textit{i.e.}, $\alpha$ and $\beta$, we first performed the grid search in a coarse level within a wide range using an adaptive step size. 
Once we obtained the approximate scope of each parameter, we then performed the fine tuning within a 
narrow range using a small step size.
And the optimal performance can be achieved when $\alpha{=}\beta{=}1$.
In addition, we
empirically set the batch-size to $32$ and the maximum number
of iterations as $500$ to ensure the convergence.

\begin{table*}[htp]
	\centering
    \scriptsize
	\setlength{\tabcolsep}{0mm}{
\begin{tabular}{p{1.4cm}<{\centering}|*{1}{p{0.9cm}<{\centering}p{0.9cm}<{\centering}
p{0.9cm}<{\centering}
p{0.9cm}<{\centering}|
p{0.9cm}<{\centering}
p{0.9cm}<{\centering}
p{0.9cm}<{\centering}
p{0.9cm}<{\centering}|}*{1}{
p{0.9cm}<{\centering}
p{0.9cm}<{\centering}p{0.9cm}<{\centering}p{0.9cm}<{\centering}|p{0.9cm}<{\centering}p{0.9cm}<{\centering}p{0.9cm}<{\centering}p{0.9cm}<{\centering}}}	
\hline	 \multirow{3}{*}{Method}&\multicolumn{8}{p{7.2cm}<{\centering}|}{MIRFLICKR-25K}&\multicolumn{8}{p{7.2cm}<{\centering}}{NUS-WIDE}\\
\cline{2-17}
&\multicolumn{4}{p{3.6cm}<{\centering}|}{Image$\rightarrow$Text}&\multicolumn{4}{c|}{Text$\rightarrow$Image}&\multicolumn{4}{p{3.6cm}<{\centering}|}{Image$\rightarrow$Text}&\multicolumn{4}{c}{Text$\rightarrow$Image}\\
\cline{2-17}
&16bits&32bits&64bits&128bits&16bits&32bits&64bits&128bits&16bits&32bits&64bits&128bits&16bits&32bits&64bits&128bits
			\\\hline 
CCA&0.621&0.602&0.586&0.573       &0.622&0.603&0.587&0.573         &0.389&0.376&0.358&0.337    &0.421&0.395&0.368&0.345\\
SCM-Or&0.632&0.588&0.564&0.552    &0.635&0.590&0.564&0.551        &0.371&0.330&0.322&0.318     &0.372&0.327&0.308&0.301\\
SCM-Se&0.738&0.750&0.761&0.765    &0.744&0.756&0.766&0.771     &0.567&0.601&0.591&0.588       &0.637&0.656&0.659&0.661\\

DCH&0.772&0.776&0.793&0.807    &0.659&0.662&0.674&0.681      &0.654&0.670&0.686&0.690    &0.584&0.622&0.640&0.639\\
DCMH&0.730&0.741&0.748&0.726    &0.759&0.767&0.775&0.749
&0.586&0.574&0.582&0.610   &0.598&0.603&0.601&0.614
\\
SSAH&0.776&0.787&0.799&0.776   &$0.773$&$0.784$&0.784&0.728      &0.615&0.616&0.618&0.529    &0.594&0.605&0.612&0.531\\
TECH&0.744&0.769&0.778&0.780   &$0.764$&$0.796$&0.805&0.805      &$\mathbf{0.674}$&0.675&$\mathbf{0.696}$&0.693    &0.706&0.719&0.725&0.733\\
SODA(ours)&$\mathbf{0.815}$&$\mathbf{0.831}$&$\mathbf{0.844}$&$\mathbf{0.847}$    &$\mathbf{0.799}$&$\mathbf{0.811}$&$\mathbf{0.822}$&$\mathbf{0.825}$    &$0.667$&$\mathbf{0.685}$&$0.695$&$\mathbf{0.702}$&$\mathbf{0.744}$&$\mathbf{0.744}$&$\mathbf{0.748}$&$\mathbf{0.763}$\\
			\hline	
		\end{tabular}
	}
   \setlength{\abovecaptionskip}{2pt}
 \caption{
	The MAP performance comparison between our proposed model and the state-of-the-art baselines on two datasets. The CLIP features are utilized for shallow learning models, and the best results are highlighted in bold.
	}\label{map1}
		\vspace{-0.9em}
\end{table*}

\begin{table*}[htp]\scriptsize
	\centering
	\setlength{\tabcolsep}{0mm}{
		\begin{tabular}{p{1.4cm}<{\centering}|*{1}{p{0.9cm}<{\centering}p{0.9cm}<{\centering}
p{0.9cm}<{\centering}
p{0.9cm}<{\centering}|
p{0.9cm}<{\centering}
p{0.9cm}<{\centering}
p{0.9cm}<{\centering}
p{0.9cm}<{\centering}|}*{1}{
p{0.9cm}<{\centering}
p{0.9cm}<{\centering}p{0.9cm}<{\centering}p{0.9cm}<{\centering}|p{0.9cm}<{\centering}p{0.9cm}<{\centering}p{0.9cm}<{\centering}p{0.9cm}<{\centering}}}	
\hline	 \multirow{3}{*}{Method}&\multicolumn{8}{p{7.2cm}<{\centering}|}{MIRFLICKR-25K}&\multicolumn{8}{p{7.2cm}<{\centering}}{NUS-WIDE}\\
\cline{2-17}
&\multicolumn{4}{p{3.6cm}<{\centering}|}{Image$\rightarrow$Text}&\multicolumn{4}{c|}{Text$\rightarrow$Image}&\multicolumn{4}{p{3.6cm}<{\centering}|}{Image$\rightarrow$Text}&\multicolumn{4}{c}{Text$\rightarrow$Image}\\
\cline{2-17}
&16bits&32bits&64bits&128bits&16bits&32bits&64bits&128bits&16bits&32bits&64bits&128bits&16bits&32bits&64bits&128bits
			\\\hline 
			CCA&0.553&0.545&0.548&0.547       &0.554&0.583&0.549&0.548         &0.306&0.299&0.294&0.290    &0.301&0.295&0.290&0.287\\
			SCM-Or&0.594&0.580&0.572&0.560    &0.605&0.590&0.567&0.555        &0.330&0.311&0.300&0.289     &0.313&0.298&0.286&0.281\\
			SCM-Se&0.686&0.691&0.691&0.694    &0.698&0.727&0.713&0.716     &0.428&0.434&0.442&0.449       &0.362&0.364&0.362&0.363\\
     
			DCH&0.638&0.642&0.662&0.669    &0.636&0.643&0.659&0.638      &0.331&0.330&0.339&0.347    &0.397&0.399&0.419&0.424\\
   DCMH&0.730&0.741&0.748&0.726    &0.759&0.767&0.775&0.749
   &0.586&0.574&0.582&0.610   &0.598&0.603&0.601&0.614
   \\
			SSAH&0.776&0.787&0.799&0.776   &$0.773$&$0.784$&0.784&0.728      &0.615&0.616&0.618&0.529    &0.594&0.605&0.612&0.531\\
 TECH&0.678&0.716&0.737&0.746   &$0.696$&$0.729$&0.747&0.754      &0.628&0.605&0.649&0.684    &0.343&0.337&0.342&0.345\\
SODA(ours)&$\mathbf{0.815}$&$\mathbf{0.831}$&$\mathbf{0.844}$&$\mathbf{0.847}$    &$\mathbf{0.799}$&$\mathbf{0.811}$&$\mathbf{0.821}$&$\mathbf{0.825}$    &$\mathbf{0.667}$&$\mathbf{0.685}$&$\mathbf{0.695}$&$\mathbf{0.702}$  &$\mathbf{0.744}$&$\mathbf{0.744}$&$\mathbf{0.748}$&$\mathbf{0.763}$\\
			\hline	
		\end{tabular}
	}  \setlength{\abovecaptionskip}{2pt}
 \caption{
	The MAP performance comparison between our proposed model and the state-of-the-art baselines on two datasets. The CNN-F features are utilized for shallow learning models, and the best results are highlighted in bold.
	}\label{map2}
		\vspace{-0.5cm}
\end{table*}

\subsection{Performance Comparison}
To justify our proposed SODA, we first compared it with baseline methods by setting four different lengths of hash codes (\textit{i.e.}, 16, 32, 64, and 128 bits) on two datasets.
Tabs.~\ref{map1} and~\ref{map2} 
show the performance comparison w.r.t. MAP among different methods.
By jointly analyzing them, we can draw the following observations:
(i) Our SODA consistently outperforms all other baselines with different hash code lengths on MIRFLICKR-25K dataset.
In particular, with the best baseline, SADA
 achieves the significant average improvement of $4.225\%$, $1.92\%$, $0.275\%$
 and $2.9\% $ in both tasks of ``Image→Text" and ``Text→Image" on
 MIRFLICKR-25K and NUS-WIDE, respectively.
This implies the advantage of our proposed cross-modal teacher-student model.
This can be attributed to the fact that, compared with the text modality, the label modality is more discriminative and is more effective to capture the semantic similarity among image instances
by mapping them into a common Hamming space.
In a sense, compared with the traditional methods that optimize hash code learning model of image using text modality directly,
the negative effect caused by the diversity of the text modality
can be avoided.
Thereafter,
the complex and diverse text modality can be optimized by fitting the pre-learned image Hamming space. 
(ii) 
Overall, the performance of SODA is significantly
 better than all baselines, except for the TECH on NUS-WIDE with the hash
 code length of $16$. Besides, when the hash code length is set as $64$, we obtain a comparable result compared with TECH.
(iii) Overall, for shallow baseline methods, the performance with the CLIP features is better than that of the CNN-F features, reflecting the strong representation capacity and the advantages of the pre-trained CLIP model.

\begin{figure*}[htp]
	\vspace{-1em}
	\centering
\begin{subfigure}{0.24\linewidth}
    \includegraphics[scale=0.18]{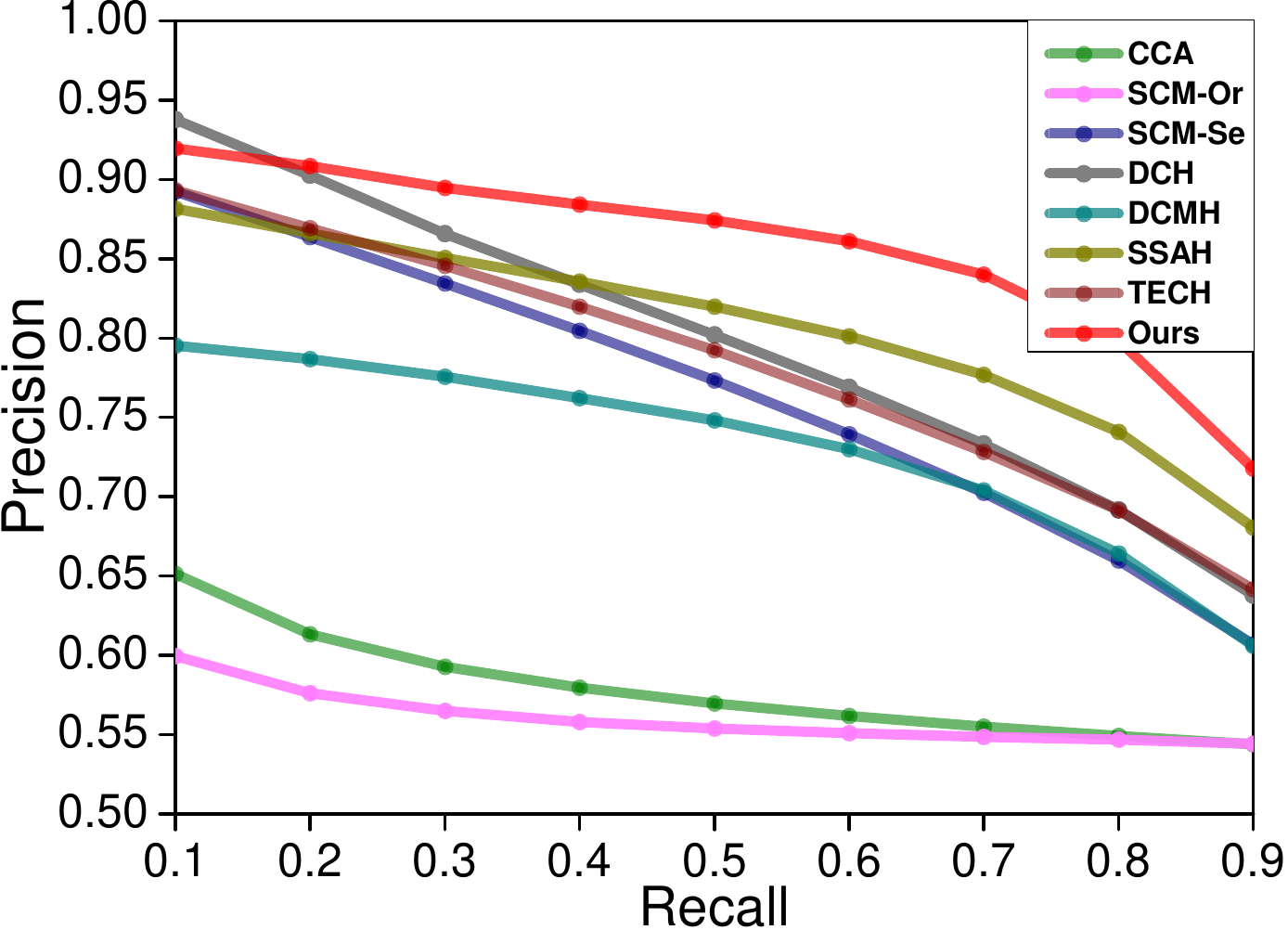}
    \caption{Image$\rightarrow$Text @ MIR25K-CLIP}
    \label{consistent}
  \end{subfigure}
\begin{subfigure}{0.24\linewidth}
    \includegraphics[scale=0.18]{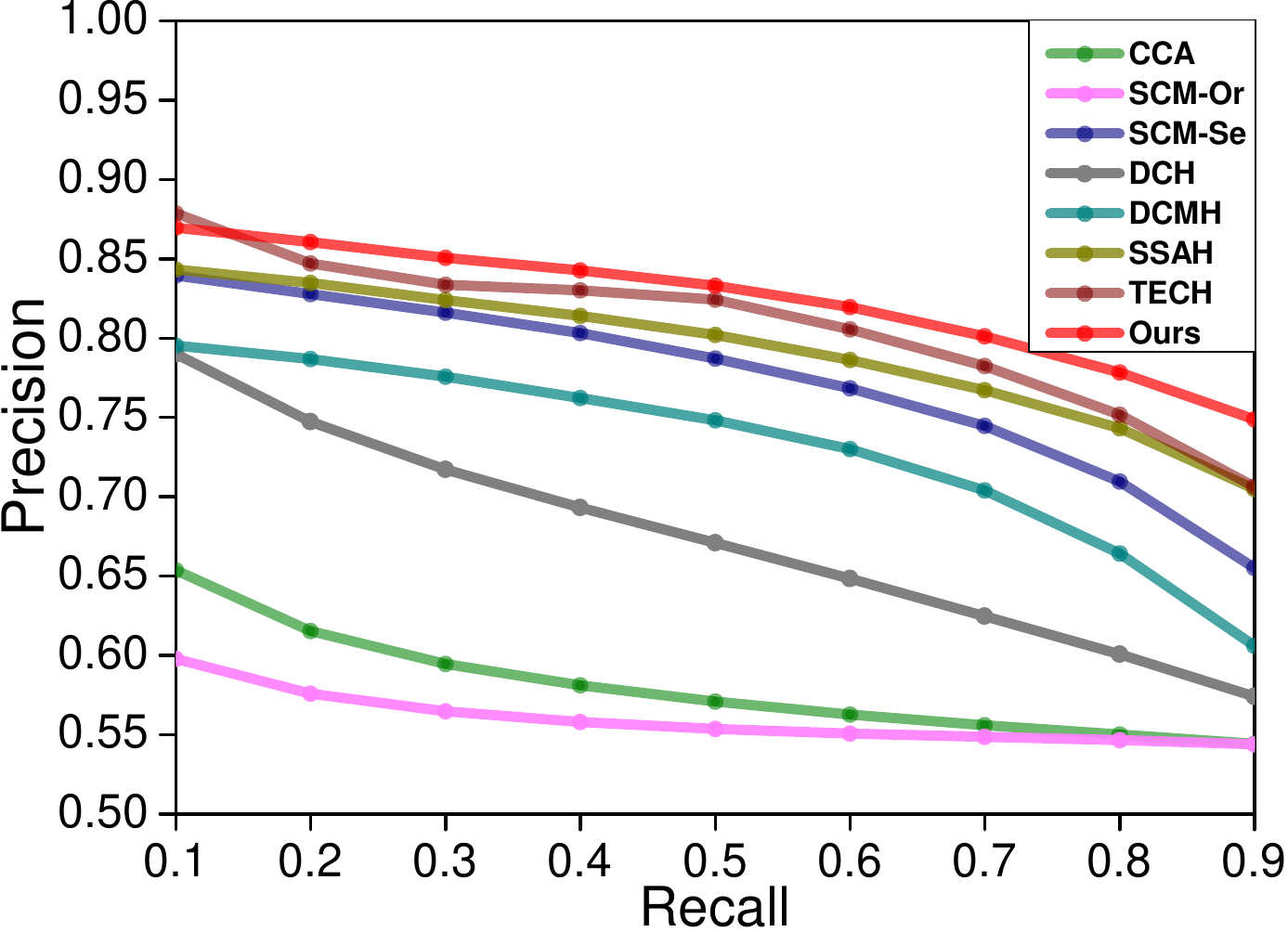}
    \caption{Text$\rightarrow$Image @ MIR25K-CLIP}
    \label{consistent}
  \end{subfigure}
\begin{subfigure}{0.24\linewidth}   
    \includegraphics[scale=0.18]{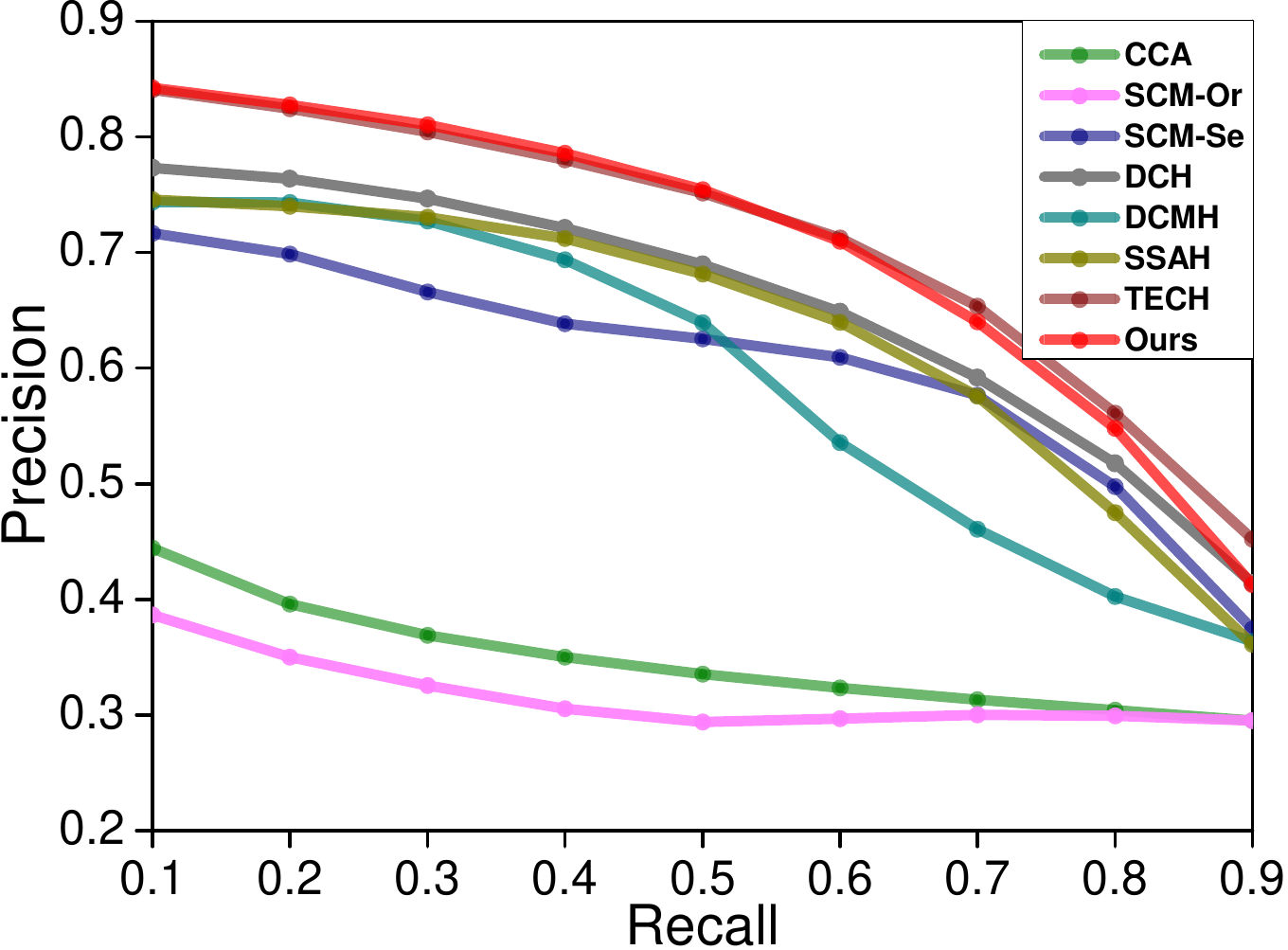}
    \caption{Image$\rightarrow$Text @ NUS-CLIP}
    \label{consistent}
  \end{subfigure}
  \begin{subfigure}{0.24\linewidth}
    \includegraphics[scale=0.18]{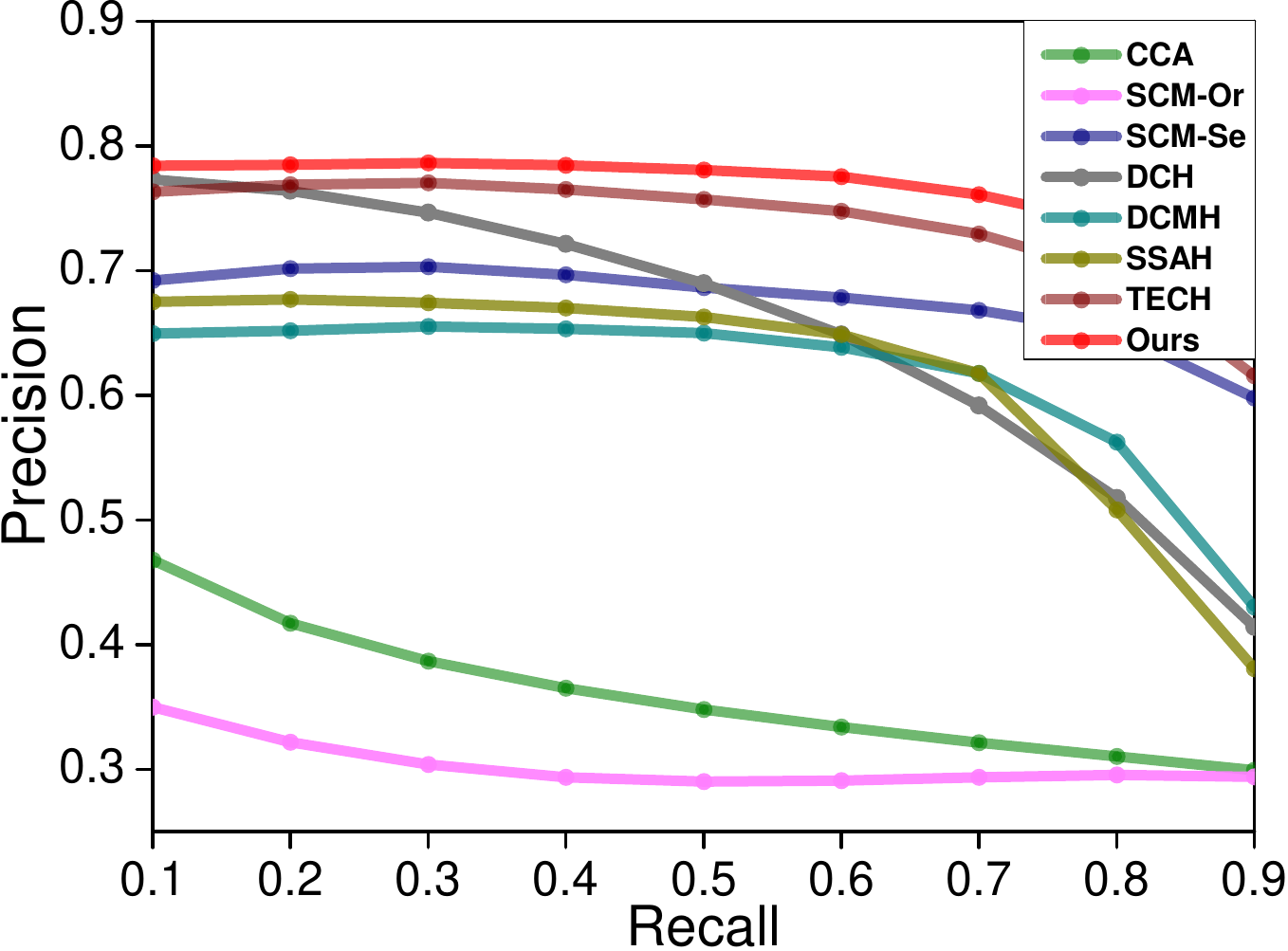}
    \caption{Text$\rightarrow$Image @ NUS-CLIP}
    \label{consistent}
  \end{subfigure}
		\setlength{\abovecaptionskip}{0.6cm}
	\vspace{-1.5em}
	\caption{ The P-R curves of different methods on two datasets, where CLIP features are utilized for baseline methods and the hash code length is $64$ bits.		
}\label{pr-curve}
\vspace{-0.25in}
\end{figure*}

To gain more deep insight, 
we further investigated the
performance of the proposed SODA on two datasets using the P-R curve with 
64 bits hash codes.
Here we chose CLIP features for shallow learning methods 
due to the fact that it brings overall more satisfactory performance compared with CNN-F features.
Specifically, we calculated the precision of returned retrieval results given different recall rate, ranging from $0.1$ to $0.9$ with a step size of $0.1$.  
As can be seen from Fig.~\ref{pr-curve},
 our SODA generally shows superiority over baselines on both datasets and has higher P-R curves, except for the situation that we obtain a comparative results of the ``Image→Text" task on NUS-WIDE dataset,
 which is 
consistent with the results in Tabs.~\ref{map1} and~\ref{map2}. 
This sheds light on the importance of devising the suitable multi-label supervision strategy to narrow the modality gap.

\begin{figure*}[htp]
	\centering
 \begin{subfigure}{0.24\linewidth}
    \includegraphics[scale=0.17]{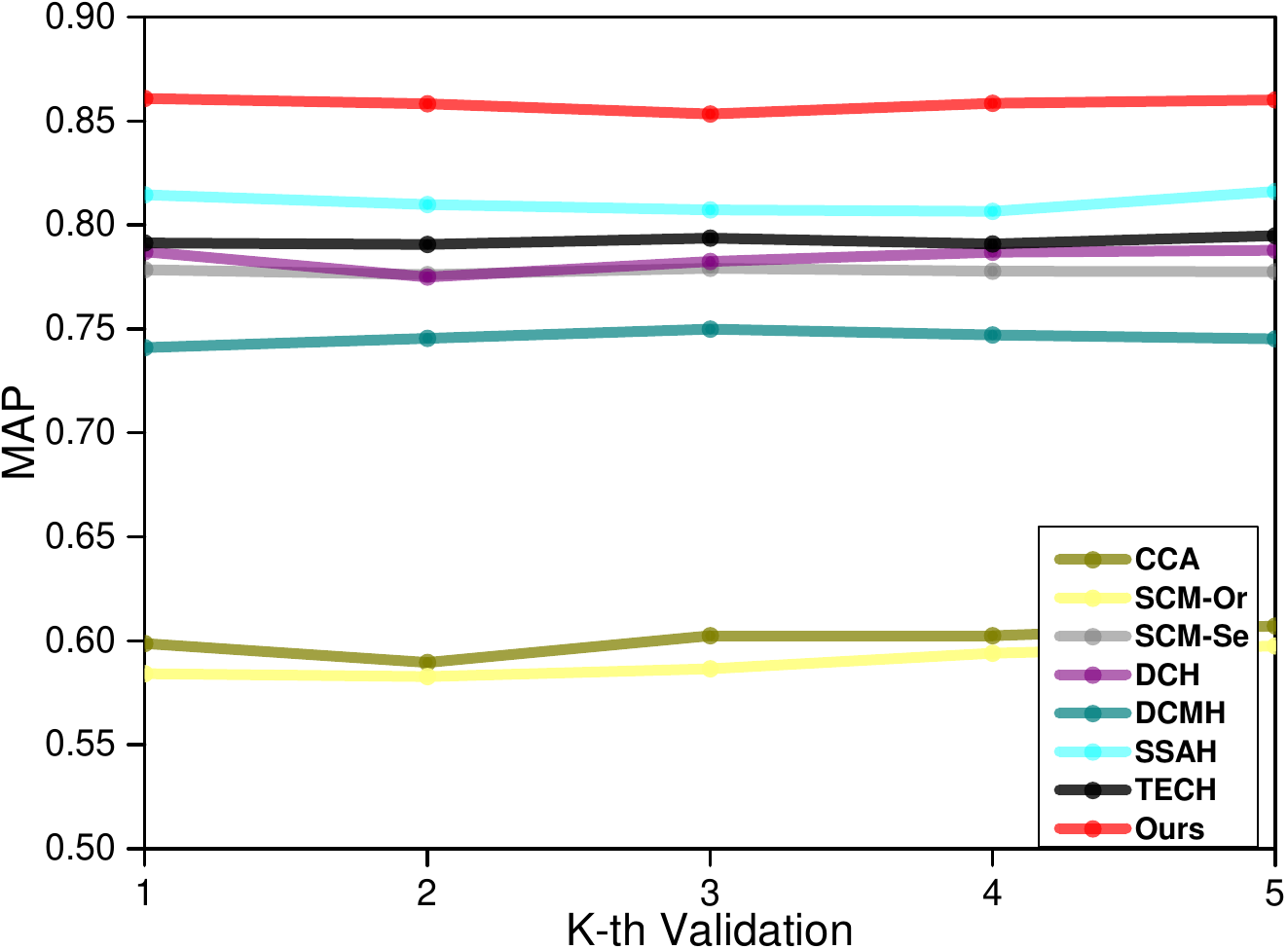}
    \caption{Image$\rightarrow$Text @  MIR25K}
  \end{subfigure}
   \begin{subfigure}{0.24\linewidth}
    \includegraphics[scale=0.17]{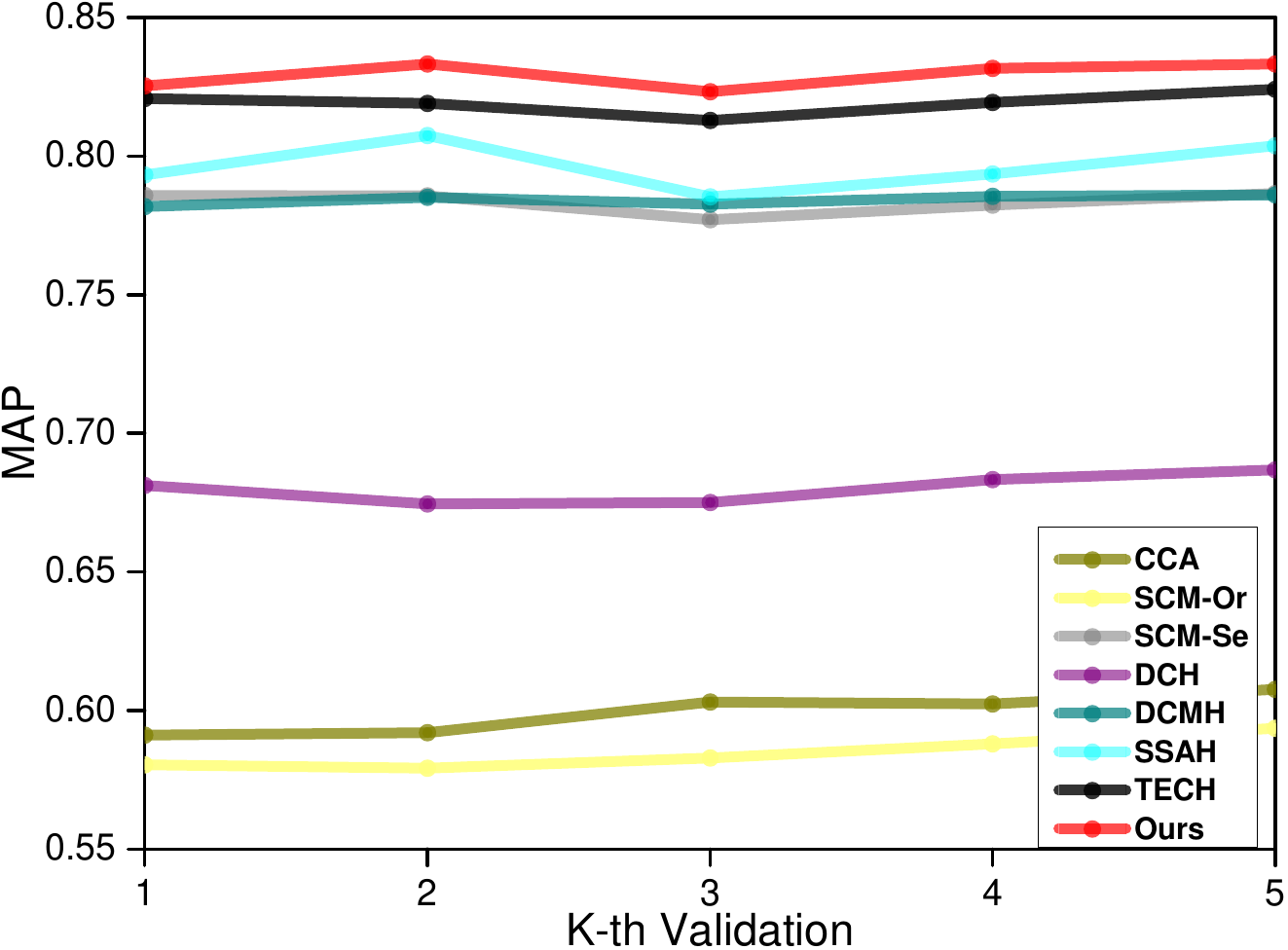}
    \caption{Text$\rightarrow$Image @ MIR25K}
  \end{subfigure}
   \begin{subfigure}{0.24\linewidth}
    \includegraphics[scale=0.17]{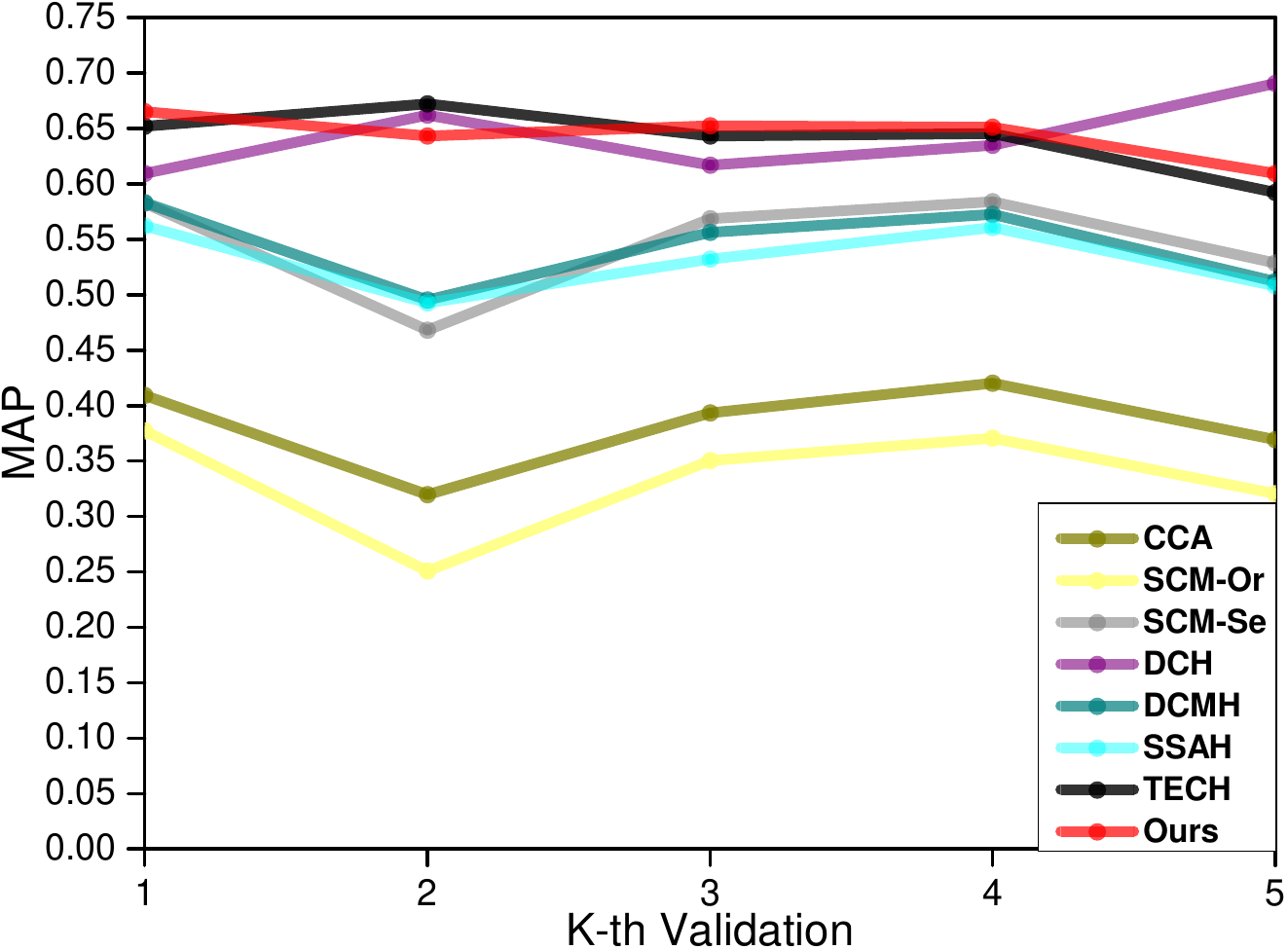}
    \caption{Image$\rightarrow$Text @ NUS}
  \end{subfigure}
     \begin{subfigure}{0.24\linewidth}
    \includegraphics[scale=0.17]{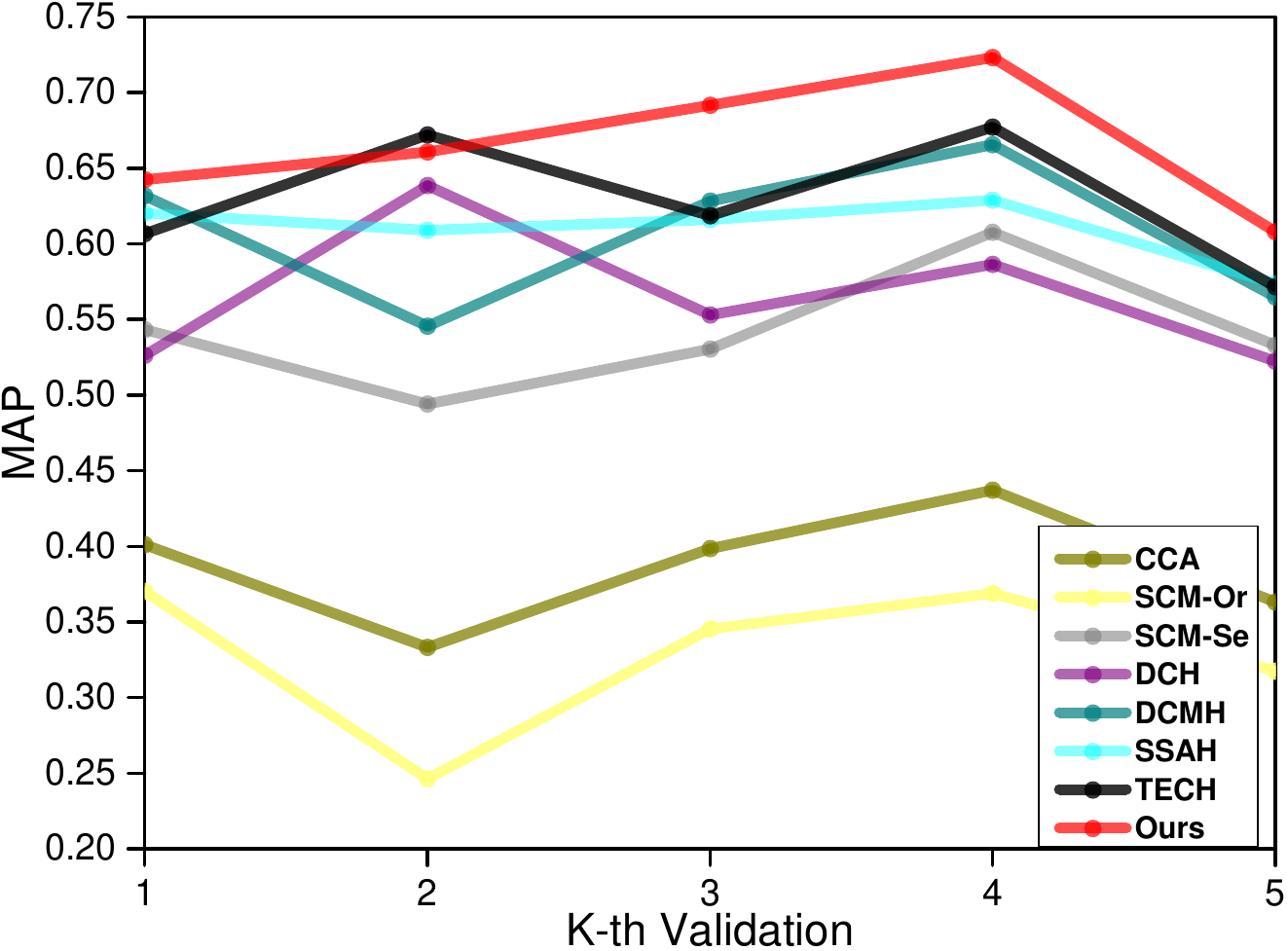}
    \caption{Text$\rightarrow$Image @ NUS}
  \end{subfigure}
			\setlength{\abovecaptionskip}{0.6cm}
	\vspace{-1.5em}
	\caption{	The five-fold cross-validation results of SODA and baseline methods on two datasets and the code length is set as $64$.
}\label{cross}
\vspace{-0.2in}
\end{figure*}

\subsection{Robustness Analysis}
To verify that our model cannot be affected by the way of dataset partition, we resorted to the idea of cross-validation. 
From Tab.~\ref{datasets}, we observed that the number of training set is quintuple than that of query set on two datasets. 
Therefore,
we performed five-fold cross-validation on two datasets, where the previously used training set are randomly divided into five equal parts,
and each part is taken in turn as the new query set, while the old query set and the remaining of training set are recombined as the new training set.
Meanwhile,
apart from instances of the new query set,
all the rest are used as gallery set.
Besides, the hash code length is set as $64$ and CLIP features are adopted.
The corresponding five-fold cross-validation results are reported
in Fig.~\ref{cross}. 
As can be seen, 
the performance is consistent with the results in Tabs.~\ref{map1} and~\ref{map2},
revealing that the superiority of our model is not random and has a good generalization and adaptability ability to fresh data.


 \begin{wraptable}{r}{0.5\textwidth}
\vspace{-0.14in} 
\centering
   \scalebox{0.7}{
  \setlength{\tabcolsep}{0mm}{
\begin{tabular}
    {p{1.5cm}<{\centering}|*{1}
  {p{1.2cm}<{\centering}
  p{1.2cm}<{\centering}
  p{1.2cm}<{\centering}
p{1.2cm}<{\centering}|
 p{1.2cm}<{\centering}
p{1.2cm}<{\centering}
p{1.2cm}<{\centering}
 p{1.2cm}<{\centering}}
  }	
  			\hline	 
   \multirow{2}{*}{Method}
   &\multicolumn{4}{p{4.8cm}<{\centering}|}{Image$\rightarrow$Text}
      &\multicolumn{4}{p{4.8cm}<{\centering}}{Text$\rightarrow$Image}   \\
			\cline{2-9}
&16bits&32bits
&64bits&128bits
&16bits&32bits
&64bits&128bits
\\
   \hline 
      DSCMR&$0.789$&$0.795$&$0.801$&$0.801$ &$0.773$&$0.782$&$0.790$&$0.802$   \\
C3CMR&$0.787$&$0.789$&$0.793$&$0.793$    &$\mathbf{0.801}$&$0.803$&$0.807$&$0.808$   \\
\textbf{SODA}&$\mathbf{0.815}$&$\mathbf{0.831}$&$\mathbf{0.844}$&$\mathbf{0.847}$    &$0.799$&$\mathbf{0.811}$&$\mathbf{0.822}$&$\mathbf{0.825}$   \\
			\hline
		\end{tabular}
	}
} 
 \vspace{-0.05in} 
\caption{Performance comparison with real-value retrieval methods on MIRFLICKR-25K.}\label{realvalue}
\vspace{-0.25in} 
\end{wraptable}
\subsection{Comparison with Real-value Retrieval}
 Apart from retrieval speed and storage cost, retrieval accuracy is also an top priority.
 Intuitively, it is inevitable that the binarization procedure will reduce retrieval accuracy.
Therefore, it is vital to ensure that the cross-modal hashing retrieval performance is comparable with real-value retrieval methods.
 To further show that the performance of our model is acceptable compared with real-value retrieval methods, we chose two classic methods DSCMR~\citep{ZhenHWP19} and C3CMR~\citep{WangGZSY22} in the real-value cross-modal retrieval field. 
As shown in Tab.~\ref{realvalue}, we reported the MAP scores of two cross-modal retrieval tasks on MIRFLICKR-25K dataset.
Apparently, for the ``Image$\rightarrow$Text" task, our proposed method 
achieves the significant improvement of all code lengths with an average value of $3.775\%$.
Besides, for the task of ``Text$\rightarrow$Image", 
our proposed method consistently surpasses real-value baselines
except for the case that SODA obtains a 
comparable and acceptable result when the code length is set as $16$.


\subsection{Parameter Sensitivity Analysis}
To check our model's sensitivity towards the core hyperparameter $\alpha$ and $\beta$ in Eqs.~\ref{alpha} and~\ref{beta}, 
we varied $\alpha$ and $\beta$ from $0.1$ 
to $1$ with a step of $0.1$ simultaneously and show the retrieval performance of two tasks on MIRFLICKR-25K 
\begin{wrapfigure}{r}{0.43\textwidth}
\centering
\vspace*{-5mm}
\includegraphics[width=1.0\linewidth]{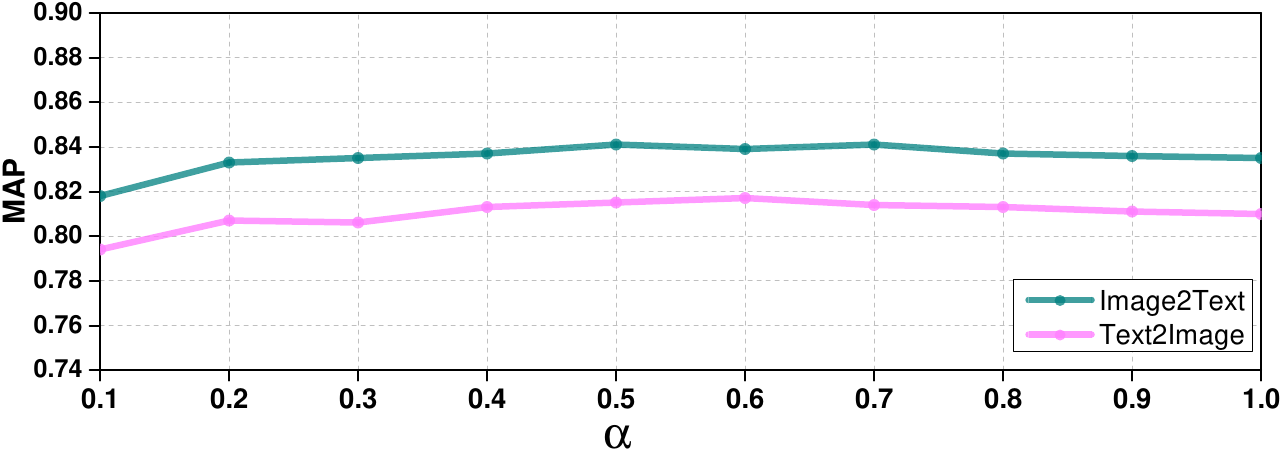}
\vspace{-0.25in}
\caption{
Sensitivity analysis of the hyper-parameters.
}
   \label{para}
\vspace*{-10mm}
\end{wrapfigure}with $64$ bit hash codes.
From Fig.~\ref{para}, 
we noticed that the performance is getting better 
with the increasing of $\alpha$.
And
the optimal performance can be achieved when $\alpha$ equals to $0.5$, indicating that both 
cross-modal hashing regularization component and binarization difference penalizing component are essential to SODA and their contributions are comparable.
Thereafter, the performance has a slight downtrend with the increasing of $\alpha$.

 \begin{wraptable}{r}{0.5\textwidth}
\centering
   \scalebox{0.7}{
  \setlength{\tabcolsep}{0mm}{
		\begin{tabular}
    {p{1.5cm}<{\centering}|*{1}
  {p{1.2cm}<{\centering}
  p{1.2cm}<{\centering}
  p{1.2cm}<{\centering}
p{1.2cm}<{\centering}|
 p{1.2cm}<{\centering}
p{1.2cm}<{\centering}
p{1.2cm}<{\centering}
 p{1.2cm}<{\centering}}
  }	
  			\hline	 
   \multirow{2}{*}{Method}
   &\multicolumn{4}{p{4.8cm}<{\centering}|}{Image$\rightarrow$Text}
      &\multicolumn{4}{p{4.8cm}<{\centering}}{Text$\rightarrow$Image}   \\
			\cline{2-9}
&16bits&32bits
&64bits&128bits
&16bits&32bits
&64bits&128bits
\\
   \hline 
SODA-\textit{it}&$0.794$&$0.818$&$0.830$&$0.835$    &$0.773$&$0.782$&$0.797$&$0.810$   \\
\textbf{SODA}&$\mathbf{0.815}$&$\mathbf{0.831}$&$\mathbf{0.844}$&$\mathbf{0.847}$    &$\mathbf{0.799}$&$\mathbf{0.811}$&$\mathbf{0.822}$&$\mathbf{0.825}$   \\
			\hline
		\end{tabular}
	}
} 
\caption{Performance of SODA and SODA-\textit{it} on MIRFLICKR-25K with different hash code lengths.}\label{ablation}
\vspace{-0.15in} 
\end{wraptable}

\subsection{Ablation Study}
To verify the effectiveness of the proposed teacher-student network and 
better explain the benefit of two-stage networks, we 
conducted comparative experiments with one derivative of our model, termed as SODA-\textit{it}.
Specifically, 
we change the hash code learning of the label modality by narrowing its modality gap between image and text modalities synchronously.
Then, the learned hash codes of label modality are utilized to supervise the learning procedures of image and text modalities.
Tab.~\ref{ablation} shows the ablation study results on MIRFLICKR-25 dataset.
From this table, 
we can find that 
our proposed SODA consistently outperforms SODA-\textit{it} over different hash code lengths.
This verifies the effectiveness of idea that first
distilling effective image modality
knowledge by narrowing the modality gap between image and label modality directly, and then 
adopting the learned image Hamming space as 
the optimization goal to the text modality to thereby realize the cross-modal similarity preserving.

\begin{wrapfigure}{r}{0.53\textwidth}
\centering
\vspace*{-4mm}
\begin{subfigure}[t]{0.45\linewidth}
    \centering
    \includegraphics[width=\linewidth]{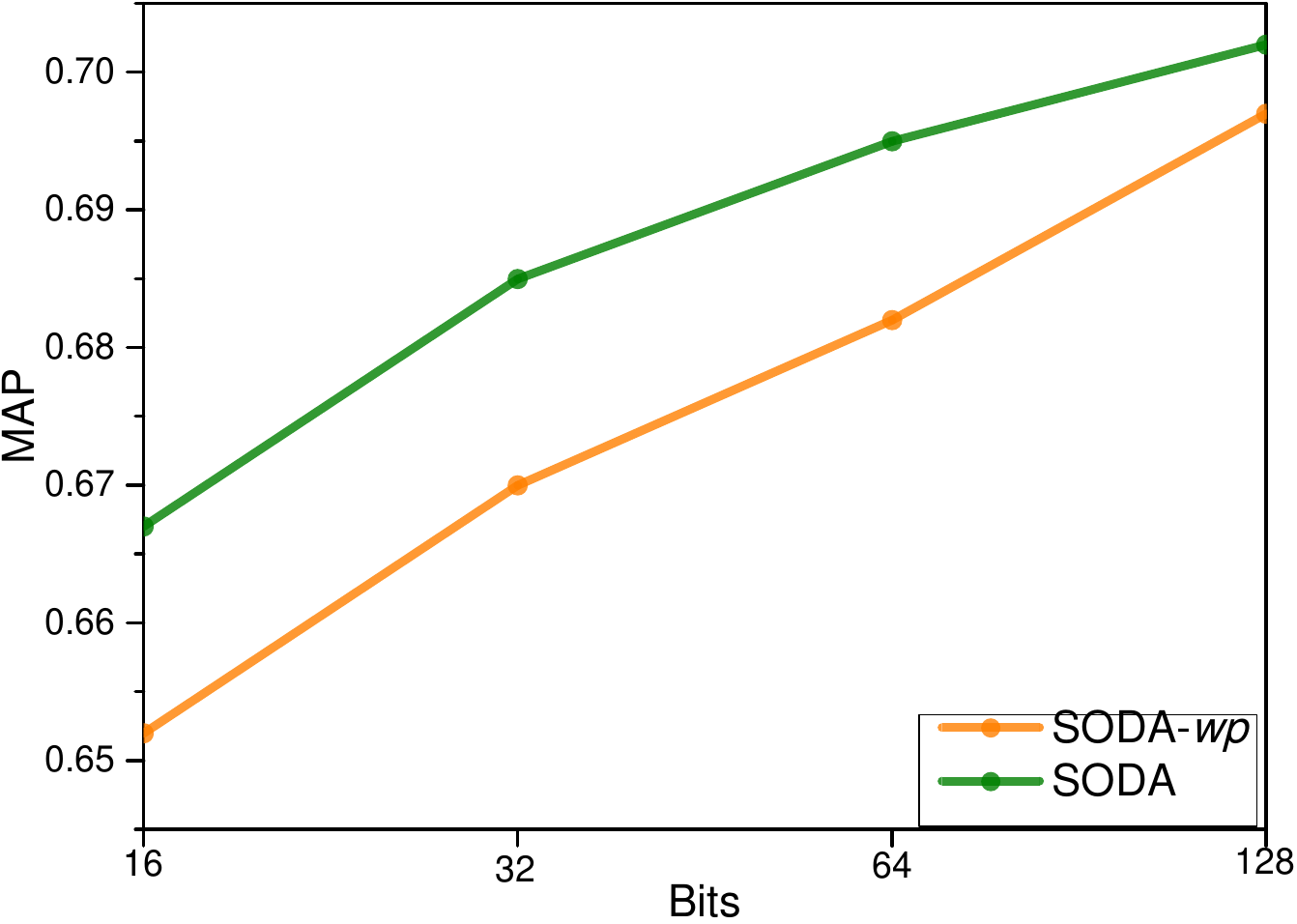}
    \caption{Image$\rightarrow$Text @  NUS}
\end{subfigure}
\hfill
\begin{subfigure}[t]{0.45\linewidth}
    \centering
    \includegraphics[width=\linewidth]{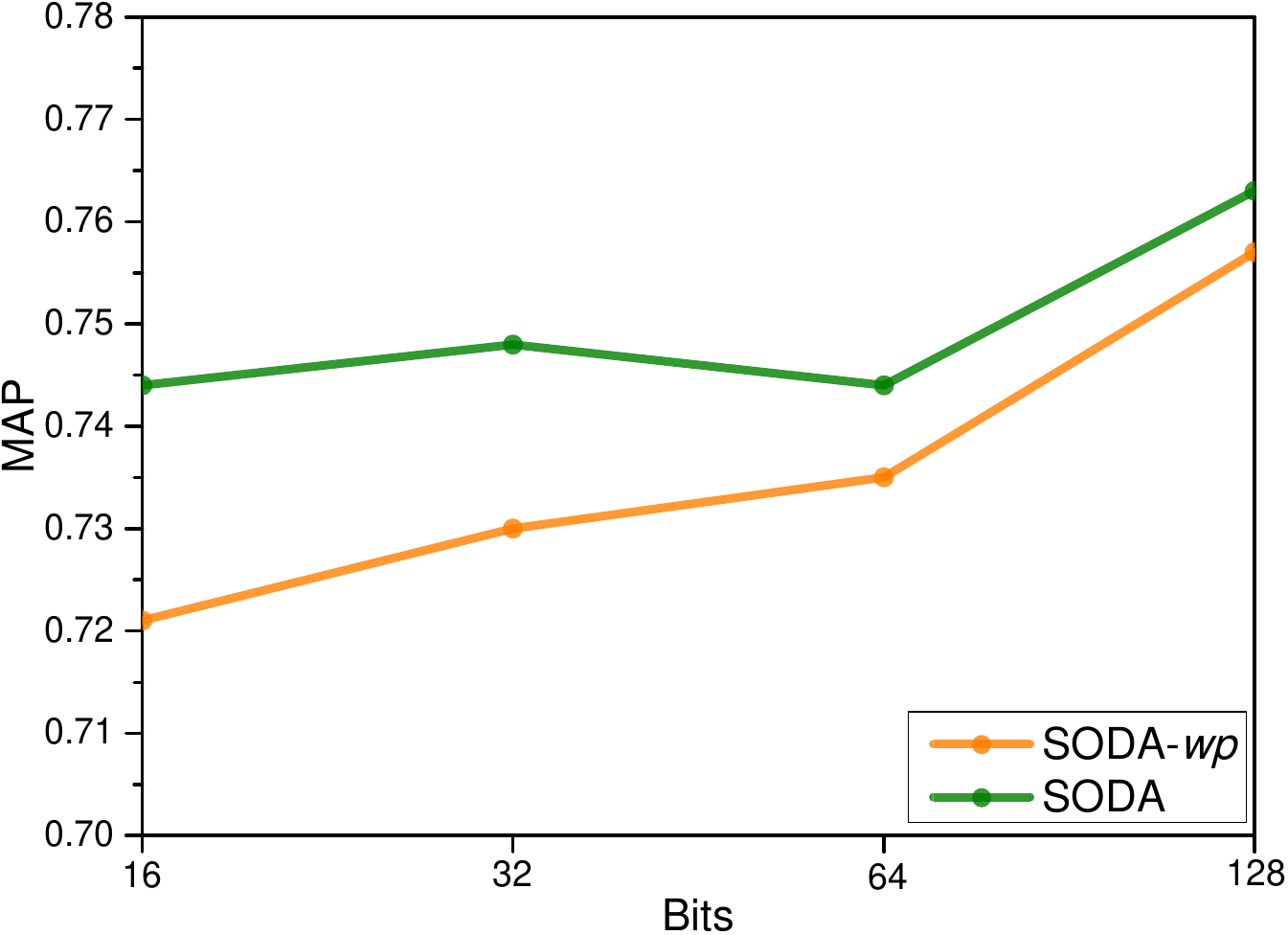}
    \caption{Text$\rightarrow$Image @ NUS}
\end{subfigure}
\vspace*{-1mm}
\caption{
Performance of SODA and SODA-\textit{wp} on NUS-WIDE with different hash code lengths.
}
\label{wp}
\vspace*{-6mm}
\end{wrapfigure}
Besides, in this work, we resorted to the 
prompt engineering~\citep{BrownMRSKDNSSAA20} and characterize 
the category labels of each instance with a set of ground-truth label prompt.
To verify its effect on our model, we conducted comparative experiments by inputting origin label description to the label encoder on NUS-WIDE and named the model as SODA-\textit{wp}. The results are demonstrated in Fig.~\ref{wp}.
As can be seen, the introduction of prompt engineering can slightly improve the performace of our proposed scheme. One possible explanation is that the prompt template make the individual label into understandable sentence, which is more acceptable for pre-trained CLIP text encoder.

%% file: 5-con.tex
\section{Conclusion}
In this paper, we focus on studying the problem of cross-model hashing
retrieval and propose a novel semantic cohesive knowledge distillation
scheme.
Compared with existing methods that adopt 
pairwise oriented and self-supervised oriented optimization strategies, we 
expect to first distill the knowledge of image modality by directly narrowing the gap between image and label modality in a cross-modal teacher network.
Then such learned image Hamming space are regarded as an optimization medium to learn the hash codes of text modality. 
Extensive experiments conducted on two real-world datasets
demonstrate the effectiveness of the proposed semantic-cohesive knowledge distillation.